\documentclass[conference]{IEEEtran}
\IEEEoverridecommandlockouts
\usepackage{cite}
\usepackage{amsmath,amssymb,amsfonts}
\usepackage{amsthm}
\usepackage{graphicx}
\usepackage{textcomp}
\usepackage{xcolor}
\usepackage{enumitem}

\newtheorem{proposition}{Proposition}

\newtheorem{problem}{Problem}

\newtheorem{remark}{Remark}

\newcommand{\R}{\mathbb{R}}
\newcommand{\x}{\mathbf{x}}
\newcommand{\g}{\mathbf{g}}
\newcommand{\M}{\mathbf{M}}
\newcommand{\bxi}{\boldsymbol{\xi}}
\newcommand{\rhat}{\hat{\mathbf{r}}}
\newcommand{\n}{\mathbf{n}}

\begin{document}

\title{Per-Agent-Safe Navigation of Globally Driven\\ Robot Collectives via Affine Field Design}


\author{Seunghwan Jang$^{1}$, Steven Ceron$^{2}$, and Wei Xiao$^{1,*}$%
\thanks{$^{1}$School of Electrical and Electronic Engineering, Nanyang Technological University, Singapore 639798. {\tt\small seunghwa001@e.ntu.edu.sg, wei.xiao@ntu.edu.sg}}%
\thanks{$^{2}$Robotics Department, University of Michigan, Ann Arbor, MI 48109, USA. {\tt\small sceron@umich.edu}}%
\thanks{$^{*}$Corresponding author.}%
}
\maketitle

\begin{abstract}
Safe navigation of robot collectives requires coordinating motion and deformation so that the collective can traverse constrained environments while ensuring each robot's safety.
Existing controllers for swarmalator collectives largely rely on local control, where each robot receives an individual command, which causes the control and communication requirements to grow with the number of agents.
We introduce a broadcast control formulation in which a single spatially affine field governs both the motion and deformation of the collective.
The scalar component of the field regulates collective size, and its spatial gradient generates collective translation through {non-reciprocal} interactions.
Both components enter the collective dynamics directly, which allows per-robot wall, actuation, and cohesion constraints to be enforced using control barrier functions within a single quadratic program for the whole collective.
The resulting controller continuously adapts the collective size to the local environment, and enables safe passage through corridors narrower than the unconstrained collective without pre-planned resizing.
We further derive how the available actuation budget determines the reachable speed and minimum passable corridor width, and we demonstrate that two independently driven collectives can simultaneously traverse a shared junction using distinct fixed phase labels without inter-collective coordination.

\end{abstract}


\section{Introduction}\label{sec:intro}

Robot collective is becoming increasingly popular as it can efficiently manage the behaviors of a large set of robot agents while performing complex tasks.
The simplest way to control a robot collective is to send the same global signal to every robot, and for sufficiently small robots this is often the only practical option
\cite{gardi2022microrobot,ceron2023programmable,xie2019reconfigurable} as they may not have local computation capability.
Broadcast control considers the setting where a large population of simple robots responds to a shared input and individual robots cannot receive separate commands \cite{becker2013massive}.
The challenge is that all of the collective behaviors, including safe navigation, must then arise from this single input.
This differs from many swarm robotics studies, where
individual robots typically execute their own programs or receive their own control commands \cite{brambilla2013swarm,rubenstein2014programmable}.

The swarmalator model \cite{okeeffe2017oscillators} offers a natural framework for controlling such collectives.
Originally introduced to describe agents that simultaneously swarm in space and synchronize an internal phase, the model can generate a wide range of collective behaviors using only a small number of global coupling parameters \cite{ceron2023diverse,okeeffe2022ring,sar2022dynamics}. These behaviors have been realized on physical robots
\cite{barcis2020sandsbots,beattie2025realizing} and have also been shown to capture the self-organization of microrobot collectives \cite{ceron2023programmable}.
More recent work has used the swarmalator model as a framework for controlling robot collectives. Non-reciprocal coupling, in which the influence of one agent on another is not necessarily equal and opposite, can generate controllable collective locomotion and formations \cite{ceron2024reciprocal,patel2026nonreciprocal}, while control barrier functions (CBFs) can be used to safely navigate these collectives through obstacle-laden environments \cite{xu2026navigation,ceron2026janus}.

Existing collective controllers, however, still depend on local control, where each robot receives its own command; for example, in \cite{ceron2024reciprocal} and \cite{patel2026nonreciprocal} each agent receives its own coupling parameter, and in the local mode of \cite{xu2026navigation} its own velocity command.
The one variant that operates under a single shared input adds a common velocity to every robot, so that the collective can only translate as a rigid body, and, as its authors report, the fixed safety margin around the formation then prevents it from passing through narrow corridors \cite{xu2026navigation}.
This limitation is not specific to one controller but follows from how the coupling has been used.
Under a shared input, prior work has regulated size through the phase coupling rather than the spatial coupling; the phase coupling enters the phase dynamics but not the position dynamics, so a barrier condition written on robot positions cannot act on phase directly, and in practice the collective is resized outside the control loop by a pre-planned transition \cite{ceron2026janus}.
However, a pre-planned resizing cannot respond to geometry that the plan did not anticipate due to uncertainties.
The alternative of placing the size parameter in a higher-order barrier \cite{ceron2024reciprocal,xiao2022hocbf} brings size back into the loop, but only at the cost of an additional integral control input.

In this paper we propose a broadcast formulation in which a single input produces motion, deformation, and safety within one optimization, by designing the input as a spatially affine field over the collective.
Each robot experiences this field at its own position, so a single control input yields a different coupling for each robot.
The scalar term in the affine field raises or lowers all couplings together and thereby sets the size of the collective, while the gradient term creates a coupling difference across the collective and, through the non-reciprocity of the interaction, drives it forward.
Both terms act on the centroid and the second moment of the collective with relative degree one, so a wall constraint written for each robot at its own position and a size bound derived from the local geometry enter a
single quadratic program (QP) as hard constraints.
The collective then compresses at every control step as a corridor narrows, without any schedule.
Beyond a single collective, assigning each collective a distinct phase label admits several of them to share a space without any inter-collective coordination.

Our contributions are as follows: (1) we formulate broadcast control of a swarmalator collective in which the control input is a spatially affine field and safety, actuation, and cohesion constraints are imposed for every agent on a single broadcast whose size does not depend on the number of agents; (2) we show that under this formulation the position and the size of the collective can both be automatically constrained by standard CBFs, so that safety and size regulation are handled in one optimization and the collective passes corridors that translation alone cannot; (3) we derive the reachable speed and the minimum passable width from a single actuation budget and verify the predicted minimum width in simulation; and (4) we show that several collectives can be operated without coordination by assigning each a distinct phase label, demonstrate two collectives crossing a junction simultaneously, and derive a bound on the number of collectives that phase labels alone can keep apart.

\section{Problem Formulation}
\subsection{The Model}\label{sub:model}
We consider $N$ robots in $D\in\{2,3\}$ dimensions whose interactions follow the swarmalator model \cite{ceron2024reciprocal}.
Robot $i$ has position $\x_i\in\mathbb{R}^D$ and phase $\theta_i\in\mathbb{R}$, and moves according to 
\begin{equation}
\dot\x_i=\frac{1}{N_0}\sum_{j}\sigma(r_{ji})
\Big[A+J_{1,i}c_{ij}-\frac{B-J_{2,i}c_{ij}}{r_{ji}}\Big]\rhat_{ji},
\label{eq:dyn}
\end{equation}
where $r_{ji}=\|\x_j-\x_i\|$ is the Euclidean distance from robot $i$ to robot $j$, $\rhat_{ji}=(\x_j-\x_i)/r_{ji}$ is the unit vector from $i$ to $j$, $c_{ij}=\cos(\theta_i-\theta_j)$ is the cosine of the phase difference, $\sigma_{ij}=\sigma(r_{ji})$ is a cutoff kernel introduced below, and $N_0$ is a normalization constant.
The coefficient $A$ scales an attraction term and $B$ is the coefficient of a repulsion term that decays as $1/r_{ji}$; these coefficients set the equilibrium spacing of the collective.
The coupling coefficients $J_{1,i}$ and $J_{2,i}$ scale the phase-dependent parts of the attraction and the repulsion, respectively.
We adopt the general form of \cite{ceron2024reciprocal}, in which the coupling enters both terms; \cite{xu2026navigation,patel2026nonreciprocal,ceron2026janus} use a repulsion that does not depend on the coupling.
Since the repulsion dominates at short range, $J_{2,i}$ carries most of the phase-dependent spatial coupling strength when the collectives are compressed.

The subscript on the coupling coefficients $J_{1,i}, J_{2,i}$ is that of the \emph{receiving} robot $i$.
The interactions are non-reciprocal \cite{fruchart2021nonreciprocal,gardi2023ondemand} because when $J_{k,i}\ne J_{k,j}, k\in\{1,2\}$, the force that robot $i$ receives from $j$ is not the negative of the force that $j$ receives from $i$, and the pair as a whole is displaced.

The second part of the swarmalator model is the phase evolution, which is defined as
\begin{equation}
\label{eqn:phase}
\dot\theta_i=\omega_i+\frac{K}{N_0}\sum_j
\frac{\sin(\theta_j-\theta_i)}{r_{ji}},
\end{equation}
where $K$ is the phase coupling gain of the Kuramoto form
\cite{kuramoto1975self} and $\omega_i$ the natural frequency. 

The original swarmalator model has no interaction cutoff, and for a single collective this is harmless, since the global attraction is what holds the collective together, but between multiple collectives this can lead to unwanted spatial attraction effects.
The repulsion in \eqref{eq:dyn} decays as $1/r_{ji}$, whereas the attraction multiplies a unit vector and does not decay at all, so two collectives can keep pulling on each other at any separation distance and never become independent.
We therefore introduce a finite-support cutoff kernel $\sigma$ \cite{lee2021finitecutoff} that confines the attraction to within a collective; beyond its support the inter-collective disturbance is exactly zero, which is what allows the per-collective reduction of Section~\ref{sec:red}.

Three assumptions frame how we design the control input, which consists of the coupling coefficients in \eqref{eq:dyn} that the broadcast sets.
First, a central controller observes the position of every robot and computes the broadcast, as in \cite{ceron2024reciprocal,xu2026navigation,ceron2026janus,patel2026nonreciprocal}.
Then the broadcast carries a constant number of values per collective, independent of $N$.
And finally, when several collectives are operated, the
command sent to one collective does not depend on the state of another; the only inter-collective coordination are the interactions due to the agents' phases.

\subsection{Global control input design}\label{sub:entry}
We design the single broadcast formulation for the collective (\ref{eq:dyn}) such that it is a spatially affine field. {Specifically, $J_{1,i}, J_{2,i}$ in (\ref{eq:dyn}) are determined by the following functions shared through all the collective: }
\begin{equation}
J_{1,i}=\lambda_1 s(\x_i),\ J_{2,i}=\lambda_2 s(\x_i), \qquad s(\x_i)=s_0+\g^\top(\x_i-\mu),
\label{eq:entry}
\end{equation}
Here, $(s_0, \g)\in \mathbb{R} \times \mathbb{R}^{D}$ is {\it the global control}, $\mu\in\mathbb{R}^D$ is the centroid of the
collective, and $\lambda_1,\lambda_2$ are the gains that convert the field value
into coupling coefficients. Each collective is driven by one such field, parameterized by the
$D{+}1$ values $(s_0,\g)$, and each robot experiences the field at its own
position. The scalar term $s_0$ raises or lowers all
couplings together and thereby sets the equilibrium size of the
collective; the gradient term $\g$ makes the coupling differ from one side
of the collective to the other and, through the non-reciprocity described
above, drives the centroid.
For simplicity, throughout the remainder of the paper we set $\lambda_1=\lambda_2=\lambda$, so that $J_{1,i}=J_{2,i}$, which we write as $J_i$, or simply $J$ when the robot index is immaterial; the absolute values of $\lambda_1, \lambda_2$ do not affect the trajectory, and the ratio $\lambda_2/\lambda_1=1$ maximizes the actuation margin, these are shown in Section~\ref{sec:limits}.

\begin{figure}[t]
\centering
\includegraphics[width=\linewidth]{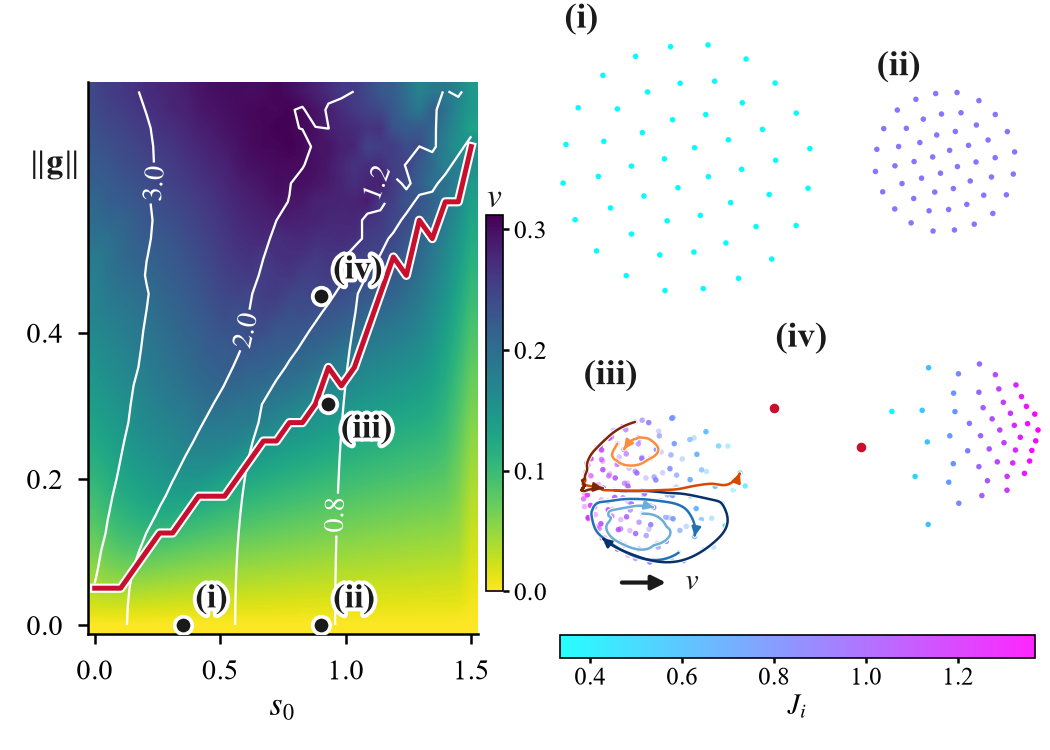}
\vspace{-24px}
\caption{The control field on the $(s_0,\|\g\|)$ plane and the collective at four points on it.
Left: color is the centroid speed $v$, white contours are the radius of the collective, and the red line is the measured boundary above which the collective fragments.
Right: the collective at pins (i)--(iv), robots colored by $J_i$.
(i), (ii) Equilibria with $\g=0$ that differ only in $s_0$.
(iii) Translation under $\g\ne0$: six robot trajectories and the direction of motion; robots circulate from front to back while the collective advances.
(iv) Beyond the boundary; red robots have detached from the main body.}
\vspace{-8px}
\label{fig:handles}
\end{figure}

The affine form is the lowest order with one term for size and one for
motion. A scalar field alone can only resize the collective; a
gradient alone can only translate it and, as Section~\ref{sec:limits}
shows, cannot pass a corridor narrower than the collective; and terms of
second or higher order act only on moments that the controller does not
use, adding broadcast values without adding a usable component.
Fig.~\ref{fig:handles} shows the two terms on the $(s_0,\|\g\|)$ plane
together with the behaviors they produce.

\begin{problem} \label{problem}
The problem we consider in this paper is to find an optimal control policy $(s_0,\g)$ for the collective (\ref{eq:dyn}) (\ref{eqn:phase}) with the affine field (\ref{eq:entry}) such that the collective can navigate in various obstacle-cluttered environments while ensuring no collisions between agents and environmental obstacles.
\end{problem}

To solve Problem \ref{problem}, we find a global control $(s_0,\g)$ for the collective employing the CBF method \cite{ames2017cbf} to ensure the safety.

\section{Collective-Level Dynamics}\label{sec:red}
The global control input $(s_0,\g)$ in \eqref{eq:entry} acts on every robot, but the
controller must regulate the speed at which the collective moves and the amount of space it occupies. This section shows that under the affine field, the dynamics of the centroid and of the second moment are affine in $(s_0,\g)$, which allows a barrier condition to be enforced by a single QP.

Let collective $a$ consist of $n_a$ robots with centroid $\mu_a$ and
second moment $\rho_a$, defined as the 
mean squared distance
of its robots from $\mu_a$ (i.e., the size of the collective), and let $\ell_1,\ell_2$ denote $\lambda_1,\lambda_2$ normalized by
$N_0$. The following proposition is derived by substituting (\ref{eq:entry}) into (\ref{eq:dyn}): 
\begin{proposition}[Collective centroid and size dynamics]\label{prop:affine}
\begin{align} \label{eqn:cent0}
\dot\mu_a&=-\M_a\g,\quad
\M_a=\frac{1}{n_aN_0}\sum_{i<j}\sigma_{ij}c_{ij}(\ell_1r_{ji}+\ell_2)
\rhat_{ji}\rhat_{ji}^\top\\
\dot\rho_a&=\Phi_a+s_0\Psi_a+\g^\top\bxi_a,\qquad \Psi_a<0,
\label{eqn:cent}
\end{align}

\end{proposition}
\noindent where $\Phi_a,\Psi_a,\bxi_a$ depend only on the current configuration (derived from (\ref{eq:dyn})).
The proposition states that the centroid moves in response to the
gradient $\g$ alone, through a gain matrix $\M_a$ assembled from the
inter-robot displacements, their directions $\rhat_{ji}$ and distances $r_{ji}$,
The second moment  in (\ref{eqn:cent}) responds to $s_0$ through
$\Psi_a$, which is negative because a uniform increase in the coupling
strengthens the attraction between every pair, so that raising $s_0$
contracts the collective. Both responses
are linear in the broadcast values. When several collectives are present,
the dynamics remain affine in $(s_0,\g)$ with additional cross terms, and
one collective's command does not enter the other collective's dynamics.

The proof of the proposition has two steps. The centroid velocity is the mean of the robot
velocities in \eqref{eq:dyn}; within each pair $(i,j)$ the $A$ and $B$
terms, which do not involve the coupling, cancel exactly because
$\rhat_{ji}=-\rhat_{ij}$ and $\sigma_{ij}=\sigma_{ji}$. The coupling terms
of a pair collect as $(J_i-J_j)\rhat_{ji}$, and because the field is
affine, $J_i-J_j=-\ell_1 r_{ji}\,\g^\top\rhat_{ji}$; each pair therefore
contributes $-\ell_1 r_{ji}\,\rhat_{ji}\rhat_{ji}^\top\g$, and the sum
over pairs is $-\M_a\g$. The same expansion applied to the second moment
separates it into a term affine in $s_0$ and a term affine in $\g$.

\begin{remark}[Validity condition]\label{rem:cohesion}
The collective dynamics (\ref{eqn:cent0}) (\ref{eqn:cent}) describes a collective only while it remains one. If the
collective fragments, the quantities $(\mu_a,\rho_a)$ no longer refer to a
single object, and the guarantees that follow apply only while cohesion is
maintained. In this work, the onset of fragmentation is identified
empirically, at a coupling ratio $J/B\approx0.5$ (Section~\ref{sec:limits}).
A closed-form prediction of the onset is left to future work. Cohesion
itself is enforced by per-robot constraints, as described in
Section~\ref{sub:rows}.
\end{remark}

In this formulation, unlike previous works, the collective-level dynamics are affine in $(s_0,\g)$, and a barrier function $h$ defined on $(\mu_a,\rho_a)$ has relative degree one with respect to the control such that a regular CBF $\dot h\ge-\gamma h$, $\gamma>0$, suffices without a high-order CBF \cite{xiao2022hocbf}.
A standard CBF \cite{ames2014cbf,ames2017cbf} can be used directly on the collective size, so $s_0$ becomes a decision variable of the same QP that enforces the wall conditions. The global controller decides how the size of the collective should evolve depending on the width of the corridor the collective is traveling through.
Previous formulations regulate the size of the collective either through $J$, which enters the position dynamics directly but has so far required a command for each robot \cite{ceron2024reciprocal,patel2026nonreciprocal}, or, under a shared input, through $K$ \cite{ceron2026janus}, whose relative degree in position is two, so that a standard CBF cannot act on it and the size is either scheduled outside the loop or placed in a high-order CBF with an additional integral input \cite{ceron2024reciprocal,xiao2019hocbf,xiao2022hocbf}.

\section{Broadcast Controller}\label{sec:ctrl}
With the collective-level dynamics in hand, the controller is a single convex QP
\cite{boyd2004convex} per collective whose $D{+}1$ decision variables are
the broadcast values; we drop the collective index $a$ where a single collective is considered.
Section~\ref{sub:rows} presents the QP
and its wall, budget, and cohesion constraints, each imposed for every
robot; Section~\ref{sub:scale} examines how the formulation scales with
the number of robots and of collectives.

\subsection{The QP and its per-robot constraints}\label{sub:rows}
At every control step the controller solves
\begin{equation}
\begin{aligned}
\min_{z=(s_0,\g)}\ &\Big(\frac{s_0-s_{0,\text{nom}}}{J_{\text{ceil}}}\Big)^2
+\Big(\frac{r_{\max}\,\|\g-\g_{\text{nom}}\|}{J_{\text{ceil}}}\Big)^2\\
\text{s.t.}\quad &\dot h_i\ge-\gamma h_i\quad\forall i
\quad\text{(wall)},\\
&|s_0+\g^\top q_i|\le J_{\text{ceil}}/\lambda\quad\forall i
\quad\text{(budget)},\\
&\dot h^{\text{coh}}_i\ge-\gamma h^{\text{coh}}_i\quad\forall i
\quad\text{(cohesion)},
\end{aligned}
\label{eq:qp}
\end{equation}
where the three constraints are defined below, $r_{\max}$ is the maximum
radius of the collective, and $q_i=\x_i-\mu$. We explain all the constraints in the following.

\textbf{Budget constraint. }The second constraint is
the actuation budget, where each agent's coupling is bounded by an
operating ceiling $J_{\text{ceil}}$, below which the collective stays
cohesive (Section~\ref{sec:limits}). Both
terms of the objective are dimensionless, so no weights are needed, and the nominal commands $s_{0,\text{nom}}, \g_{\text{nom}}$ are obtained from
Proposition~\ref{prop:affine} and the budget.
Inverting $\dot\mu=-\M\g$ for a goal direction $\hat e$ gives the direction $-\M^{\dagger}\hat e$, and $\g_{\text{nom}}$ is this direction scaled to the largest magnitude the budget allows.
The nominal $s_{0,\text{nom}}$ is the smallest value for which no agent's coupling
under $\g_{\text{nom}}$ falls below the cohesion floor $J_{\text{floor}}$,
namely $(r_{\max}\|\g_{\text{nom}}\|_1-J_{\text{floor}}/\lambda)^{+}$,
where $J_{\text{floor}}$ is the coupling below which the collective
disperses.

\textbf{Wall Constraint. }The wall constraint (i.e., collision-free from the walls) is written for each robot at its own position. For a
wall with unit normal $\n$ through a point $\mathbf{p}$, and a margin
$d_{\text{wall}}$,
\begin{align}
h_i&=\n^\top(\x_i-\mathbf{p})-d_{\text{wall}},\\
\dot h_i&=\n^\top F_i+s_0(\n^\top w_i)+\g^\top[q_i(\n^\top w_i)],
\end{align}
where $F_i$ is the part of robot $i$'s velocity that does not depend on
the control and $w_i$ the velocity robot $i$ receives per unit coupling.
Each robot contributes one linear condition in the
$D{+}1$ decision variables; each (robot, wall) pair is a half-space in
$(s_0,\g)$, their intersection is a convex polytope, and the QP projects
the nominal command onto it.

Writing the constraint per robot, rather than once for the collective,
is what allows narrow passages. The global controller of
\cite{xu2026navigation} imposes one condition per obstacle on the
collective as a whole, treating it as a ball, and it is the fixed margin
of that ball that prevents passage through narrow corridors. A per-robot
constraint measures clearance where each robot actually is, and the
difference is not small; in our experiments the gap between the ball
radius and the true extent of the collective along the wall normal
exceeds the equilibrium radius of the collective itself.

The actuation budget is likewise imposed per robot, as the second
constraint of \eqref{eq:qp}, which adds two linear rows per robot without
adding decision variables. A single
lumped bound $|s_0|+r_{\max}\|\g\|_1$ would assume worst-case alignment
and be conservative. Since the left-hand side is linear in $q_i$, its
maximum over the collective is attained at a vertex of the convex hull of
the robot positions, so imposing the constraint on the hull vertices alone
is exactly equivalent to imposing it on every robot. The reachable speed
of the collective is the maximum over this budget polytope,
Eq.~\eqref{eq:vdir} shown later.

\textbf{Cohesion constraint. } The second moment $\rho_a$ is affine in the control by
Proposition~\ref{prop:affine}, but it bounds the mean squared distance radius rather than
the maximum. A few outlying robots change $\rho_a$ very little while
inflating $r_{\max}$, and since $r_{\max}$ is the denominator of the
budget, $\|\g\|\le(J_{\text{ceil}}-|s_0|)/r_{\max}$, steering authority
drops with them. We therefore impose cohesion as a per-robot constraint in
the same form as the wall,
\begin{equation}
h^{\text{coh}}_i=r_{\text{cap}}(\x)-\|\x_i-\mu\|\ \ge\ 0,
\label{eq:coh}
\end{equation}
which has relative degree one, adds no decision variables, and reduces to
the hull vertices as the budget does. The bound $r_{\text{cap}}$
is a geometric quantity, i.e., the local free-space half-width at the current
position minus $d_{\text{wall}}$, so it requires neither a schedule nor a
tuning constant. As an opening narrows, the bound descends with it, and
the adjustment takes place inside the optimizer.

\begin{figure}[t]
\centering
\includegraphics[width=\linewidth,height=0.55\linewidth,keepaspectratio]{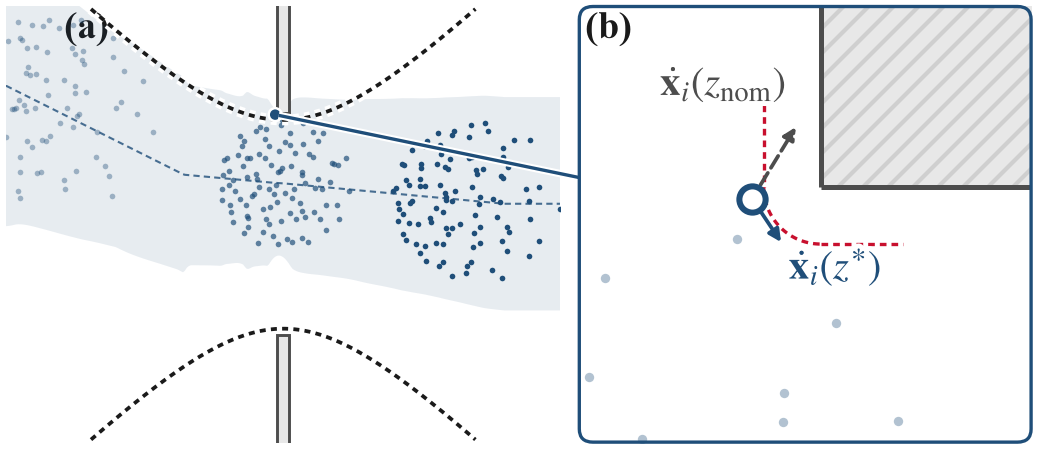}
\vspace{-16px}
\caption{Passage of the 3-D partition window, projected onto the plane normal to the window.
(a) The dashed curve is the cohesion bound $r_{\mathrm{cap}}(\x)$ set by the window, the shaded band is the measured extent of the collective, and the three snapshots are earlier when lighter.
(b) The robot marked in (a) at the instant the wall constraint
is most active: the dashed arrow $\dot\x_i(z_{\text{nom}})$ is the command the QP would issue
without the wall constraint, the solid arrow $\dot\x_i(z^*)$ the command it issues with it, and the dashed contour the safety boundary at margin
$d_{\mathrm{wall}}$.}
\vspace{-15px}
\label{fig:funnel}
\end{figure}

Collision safety is already guaranteed by the wall constraints, but those are reactive, so
without the cohesion constraint compression piles up just before an
opening and robots at the edge catch on its rim; the cohesion constraint
induces the compression in advance and yields a smooth passage
(Fig.~\ref{fig:funnel}). We impose cohesion as a constraint rather than
tracking a reference radius because a reference would compete with goal
tracking, whereas a constraint, once satisfied, leaves the remaining
budget entirely to forward progress.

\subsection{Scale in robots and in collectives}\label{sub:scale}
The decision variables of the QP are the broadcast values $s_0$ and $\g$, which consist of one scalar and one $D$-dimensional vector, for $D{+}1$ variables independent of the number of robots $N$. In per-agent control, both
the decision variables and the transmitted commands grow in proportion to
$N$, and a centroid-and-radius controller uses three. The constraints, by contrast, arise one per robot, but
they reduce to the vertices of the convex hull, and since those vertices
lie on the rim of the collective, their number grows as $\sqrt N$. In
simulation the row count grows as $N^{0.46}$ and the QP solve time as
$N^{0.84}$, both sublinear. Communication scales the same way, each cycle
broadcasts $D{+}1$ values, whereas per-agent control requires $ND$
command values and neighbor exchange $N(N{-}1)D$; at $N=400$ these
are $3$, $800$, and $319{,}200$.

With $M$ collectives, each collective contributes one scalar $s_0$ and one $D$-dimensional vector $\g$, which gives $M(D{+}1)$ decision variables. The QP of one collective involves no other collective's command, but in the
dynamics the collectives are coupled through cross terms whose sign and
magnitude are set by their phase difference
(Proposition~\ref{prop:affine}), and the remaining design choice is
whether to include those cross terms in the QP model.
\begin{proposition}[Definiteness of the gain]\label{lem:pd}
The intra-collective gain $\M_a$ of \eqref{eqn:cent0}, written as $\sum_{i<j}w_{ij}\rhat_{ji}\rhat_{ji}^\top$, has
$w_{ij}>0$ for every pair when $\cos\Delta\theta=1$ within a
collective; it is a sum of positive rank-one matrices, positive
semidefinite, and positive definite whenever the pair directions span
$\R^D$. The only structural source of indefiniteness is the cross term
contributed by pairs from different collectives, for which $\cos\Delta\theta<0$.
\end{proposition}
We exclude the cross term from the QP. By Proposition~\ref{lem:pd} the
intra-collective gain is positive definite by construction, whereas the
cross term becomes degenerate when collectives fuse. The QP model is then
an approximation whenever collectives are close, but it never degenerates,
so the controller retains a well-defined steering direction even when
collectives are in contact.

\section{Conditions and Limits of a Single Input}\label{sec:limits}
The controller of Section~\ref{sec:ctrl} enforces safety and size with one
broadcast, but it does so within limits that the broadcast imposes.
We first show that size
regulation is a necessity, since translation alone
cannot satisfy a two-sided wall constraint; we then show that the constant
and the gradient terms draw on one actuation budget, from which the
reachable speed and the narrowest passable corridor follow; and we close
with the bound that phase labels place on the number of collectives.

\begin{figure}[t]
\centering
\begin{minipage}[t]{0.48\linewidth}
\centering
\includegraphics[width=\linewidth]{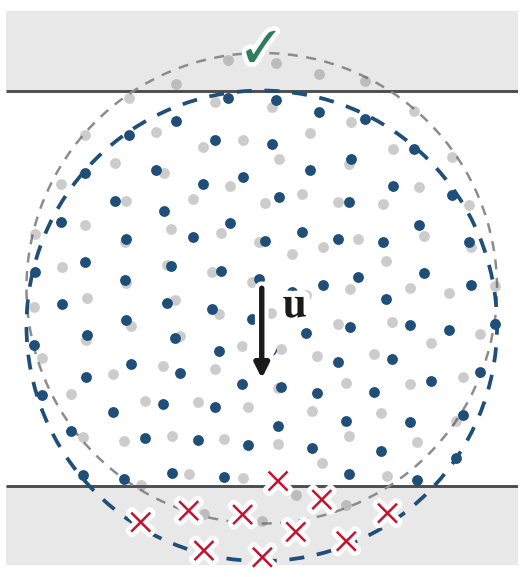}\\
{\footnotesize (a) translation only}
\end{minipage}\hfill
\begin{minipage}[t]{0.48\linewidth}
\centering
\includegraphics[width=\linewidth]{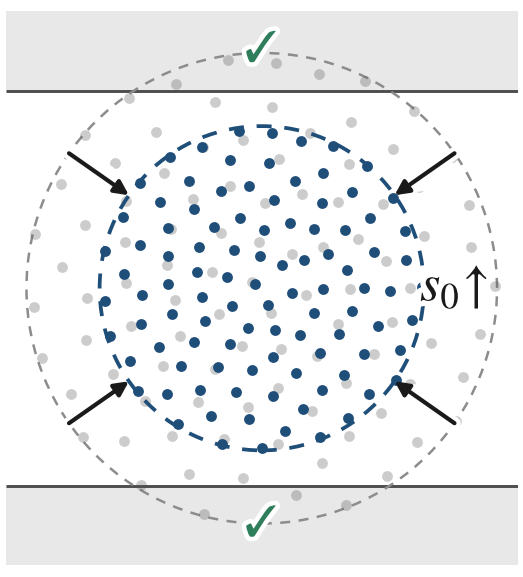}\\
{\footnotesize (b) with $s_0$}
\end{minipage}
\caption{Why $s_0$ is a necessity. (a) With translation alone the two
squeezed robots demand opposite motions and cannot both be satisfied.
(b) With $s_0$ the collective contracts and both are satisfied.}
\vspace{-12px}
\label{fig:translation}
\end{figure}

A plain broadcast translation \cite{xu2026navigation} cannot pass a corridor narrower than the
collective, for the simple geometric reason that the robots nearest the two
walls each need to move away from their wall, in opposite directions along
the wall normal $\n$, and a translation moves every robot the same way.
Formally, if the two robots require $\n^\top\mathbf{u}\ge b_1$ and
$-\n^\top\mathbf{u}\ge b_2$ of a common velocity $\mathbf{u}$, adding the
two shows that both can hold only if $b_1+b_2\le0$, that is, only if there is room between the walls.
Only a change of shape can satisfy both constraints once there is not (Fig.~\ref{fig:translation}), which places the feasibility boundary of any translation-only broadcast near a corridor-to-collective width ratio
of one.
We verified this with a linear program (LP) that tests the static feasibility of the wall constraints as the ratio is varied.
The LP places the feasibility boundary of a broadcast translation between ratios $1.0$ and $1.1$, where the global controller of \cite{xu2026navigation} operates at $1.098$ in agreement with the limit that work reports, whereas the affine field remains feasible below it, taking the largest admissible $s_0$ as the ratio approaches the boundary.

Size regulation, however, is not free. Both $s_0$ and $\g$ act on the
robots through the same coupling coefficient, which is bounded by the
ceiling $J_{\text{ceil}}$, so compression and propulsion compete for one
budget of actuation authority. Under the per-robot bound
$|J_i|\le J_{\text{ceil}}<B$ the admissible set of $(s_0,\g)$ is the
polytope $P=\{(s_0,\g):|s_0+\g^\top q_i|\le J_{\text{ceil}}/\lambda\ \forall i\}$,
and the reachable speed along a corridor axis $\n$, given the compression
$s_0^{\text{req}}$ that the corridor demands, is the linear program
\begin{equation}
v_\n=\max_{(s_0,\g)\in P,\ s_0=s_0^{\text{req}}}\ (-\n^\top\M\g).
\label{eq:vdir}
\end{equation}
The quantity that governs the trade-off is the headroom
$c=J_{\text{ceil}}/\lambda-|s_0|$, the authority that remains for propulsion once compression has taken its share.
Equation~\eqref{eq:vdir} agrees with simulation where the budget binds.
The ceiling $J_{\text{ceil}}<B$ keeps the repulsion coefficient $B-J_{2,i}\cos\Delta\theta$ positive; in simulation the collective fragments well before that, from $J/B\approx0.5$, and we take this as the operating ceiling.

The budget yields the narrowest passable corridor.
\begin{proposition}[Minimum passable width and time divergence]\label{prop:wmin}
The minimum passable width $W_{\min}$ is the width at which the
equilibrium radius at the maximum compression
$s_0^\ast=J_{\text{ceil}}/\lambda$ no longer fits the corridor, and as $W$
approaches $W_{\min}$ the traversal time diverges inversely in the
remaining headroom, $t\propto1/c(W)$, where $c(W)=J_{\text{ceil}}/\lambda-s_0^{\text{req}}(W)$ decreases as the required compression grows.
\end{proposition}
Both the width and the divergence follow from quantities that can be
computed before any traversal is simulated. The equilibrium spacing of the
collective at a given coupling follows from the interaction law as
$(B-J)/(A+J)$, and the ratio of the outermost radius to that spacing is a
shape factor that we measured once from a static relaxation and found to
be independent of the coupling; the coupling ceiling therefore fixes
$W_{\min}$. The slope of the divergence follows from the LP as
well, such that with $c/v_\n$ nearly constant across widths, the traversal time is
$L_{\text{path}}\,c/v_\n$ ($L_{\text{path}}$ is the corridor length), and
after correcting the effective path for the arrival radius this prediction
agrees with the simulated completion time to within one percent
(Fig.~\ref{fig:budget}). Although the reachable speed is a maximum over the
polytope $P$ and not a function of $c$ alone, the shape of $P$ is
fixed within one corridor family, so the law reduces to $t=a/c+b$. We
computed $W_{\min}$ from the static equilibrium and registered it before
running the corridor experiments; Section~\ref{sub:single} reports the
comparison. The prediction rises slightly with the number of robots $N$,
because the outermost radius is a maximum over $N$ samples and approaches
its limit as $N$ grows; all of our experiments in this work use $N=50$.

Near $W_{\min}$ the failure of a traversal takes one of two forms
(Table~\ref{tab:regimes}). When the corridor is only slightly too narrow,
the cohesion bound can still be met but no headroom remains for $\g$, and
the traversal time diverges as the proposition predicts; when the corridor
is narrower still, the bound itself cannot be met and the collective
stalls. Both boundaries follow from the same polytope.

The budget also determines how the coupling should be split between the
attraction and the repulsion terms, and here the absolute gain is not important. Since the field value $s$ enters \eqref{eq:dyn} only through $J=\lambda s$,
the trajectory is unchanged under $(\lambda,s)\to(\alpha\lambda,\,s/\alpha)$ once
the constraint constants $J_{\text{ceil}}/\lambda$ and $J_{\text{floor}}$
are scaled with it, and since the repulsion term is homogeneous of degree
zero under a common scaling of $B$, $\lambda_2$, and length while the
attraction does not involve the distance, the dimensionless trajectory is likewise
invariant under that similarity; both invariances are confirmed in
simulation. The overall gain therefore cancels, but the ratio
$\kappa=\lambda_2/\lambda_1$ does not, since $\M=\lambda_1\tilde\M(\kappa)$,
and the ratio has a feasible band.
\begin{proposition}[Split-ratio band]\label{prop:ratio}
With $\lambda_1$ normalized to one, so that $\lambda_2=\kappa$, the compression required at operating radius
$r$ is $s_0^{\text{req}}(r,\kappa)=(B-Ar)/(\kappa+r)$, and under the symmetric
ceiling $|J_{1,i}|,|J_{2,i}|\le J_{\text{ceil}}$ the admissible $s$ is bounded by
$J_{\text{ceil}}/\max(1,\kappa)$. The headroom
\begin{equation}
c(\kappa)=\frac{J_{\text{ceil}}}{\max(1,\kappa)}-\frac{B-Ar}{\kappa+r}
\label{eq:crho}
\end{equation}
increases for $\kappa<1$ and decreases for $\kappa>1$, is therefore maximal at
$\kappa=1$, and $c(\kappa)=0$ bounds the feasible band $(\kappa_-,\kappa_+)$.
\end{proposition}
We take the operating radius $r$ to be the largest the corridor admits.
Inside the band a ratio with small headroom still fails by time
divergence, so the feasibility verdict closes only when
Propositions~\ref{prop:wmin} and~\ref{prop:ratio} are applied together;
that combined verdict matched every point of the split-ratio sweep
reported in Section~\ref{sub:single}. Because the ratio changes the shape
of $\tilde\M$, the time coefficients fitted at $\kappa=1$ do not transfer to
other ratios.
The limits above concern a single collective. With several collectives,
the remaining limit comes from the phase labels.
\begin{proposition}[Bound on the number of phase-separated collectives]\label{prop:phase}
For the phase coupling matrix $C_{ij}=\cos(\theta_i-\theta_j)$ of $M$
collectives,
all pairs can be orthogonal ($C=I$) only if $M\le2$, all
pairs can be equally coupled only if $M\le3$, and for $M\ge4$ every
assignment contains a pair with
$\cos(\theta_i-\theta_j)\ge\cos(2\pi/M)\ge0$.
\end{proposition}
The matrix $C$ is the Gram matrix of the unit vectors
$(\cos\theta_k,\sin\theta_k)$ in the plane and has a rank of at most two,
and among any $M$ angles on the circle some two differ by at most
$2\pi/M$. Section~\ref{sub:multiF} shows what this bound means for the
number of collectives that can share a workspace.

\section{Results}\label{sec:results}

\subsection{Single collective}\label{sub:single}
We show a corridor experiment which tests whether the minimum passable width and the
time divergence predicted in Section~\ref{sec:limits} hold, by
traversing a straight corridor of varying width $W$. 
The maze experiment uses the two- and three-dimensional mazes and reference path of \cite{ceron2026janus} and compares our approach with that paper's global translation and scheduled size regulation, re-implemented on our model (the prior scheme), on geometry that this scheme cannot handle; the simulator extends the code of \cite{patel2026nonreciprocal}.
Before the experiments we verified the
reduction of Section~\ref{sec:red} numerically such that its residual is of order
$10^{-16}$, the finite-difference convergence order is two, and the
hull-vertex budget matches the full per-robot expansion exactly.

\begin{figure}[h]
\centering
\includegraphics[width=0.8\linewidth]{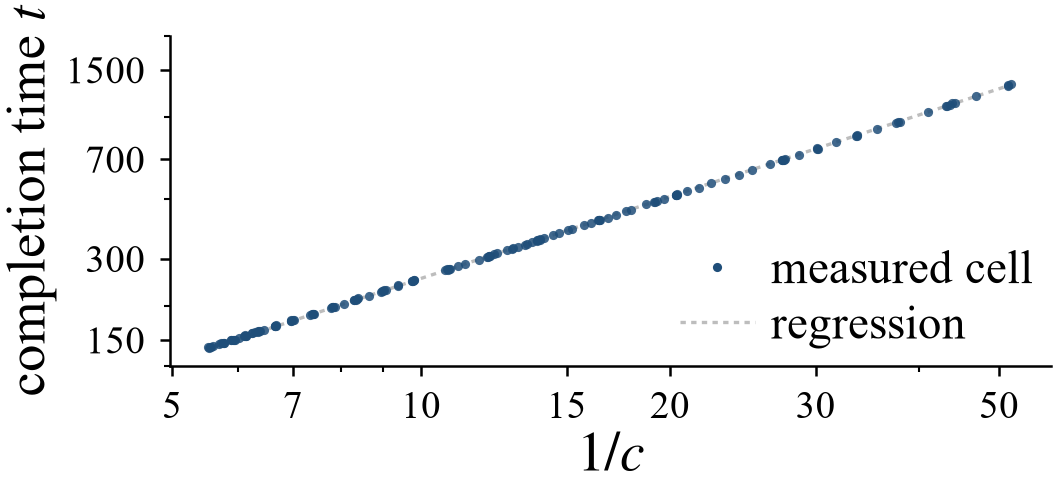}
\vspace{-8pt}
\caption{Predictive power of the actuation budget. Traversal time against inverse headroom $1/c$, with the regression fit to all cells in the passing regime.}
\label{fig:budget}
\end{figure}

\vspace{-8pt}
\begin{table}[h]
\caption{Three regimes of the corridor experiment}
\label{tab:regimes}
\centering\footnotesize
\begin{tabular}{ll}
\hline
width $W$ & outcome\\\hline
$\le1.600$ & cohesion bound unmet; stall\\
$1.620$--$1.670$ & bound met; goal not reached (time divergence)\\
$\ge1.680$ & goal reached, collision-free\\\hline
\end{tabular}
\end{table}
\vspace{-4pt}
The corridor results confirm the predictions of Section~\ref{sec:limits}.
The pre-registered minimum width $W_{\min}^{\text{pred}}=1.636$ lies
within three percent of the boundary observed in simulation,
$(1.670,1.680]$ (Table~\ref{tab:regimes}), so the narrowest passable
corridor is set by the dynamics rather than by controller tuning, as
Proposition~\ref{prop:wmin} states. The completion time follows
$t=25.58/c-2.37$ with $R^2=0.99993$, inversely in the headroom and
monotone in the width, so the form of the divergence holds as well
(Fig.~\ref{fig:budget}). A crossing check finds no wall collision at any width, and in every
passing run the global minimum of the barrier $\min_i h_i$ stays at the
solver tolerance, $-2.2\times10^{-5}$, so the wall margin is maintained
at every step. Across the sweep of the split ratio $\kappa$, the combined verdict of Propositions~\ref{prop:ratio}
and~\ref{prop:wmin} predicts pass or fail correctly in every case, and
$\kappa=1$ has the largest headroom (Fig.~\ref{fig:sweeps}(b)); in the $J/B$ sweep, fragmentation
begins at $0.50$, which is the ceiling adopted in
Section~\ref{sec:limits}.
\vspace{-4pt}

\begin{figure}[t]
\centering
\includegraphics[width=\linewidth,height=0.5\linewidth,keepaspectratio]{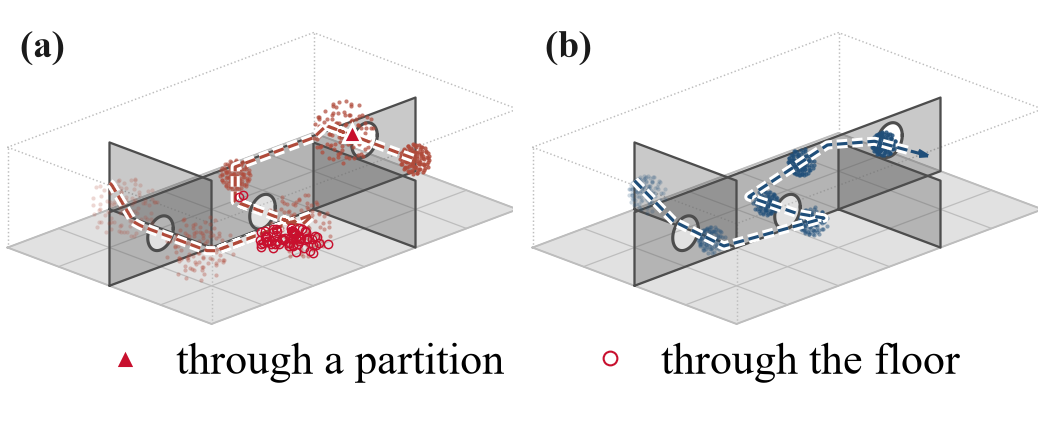}
\vspace{-24px}
\caption{The same 3-D maze under both schemes, with collision sites marked. (a) The prior scheme collides with the walls and floor. (b) Our approach is collision-free.}
\vspace{-8px}
\label{fig:maze3d}
\end{figure}

\begin{figure}[t]
\centering
\includegraphics[width=\linewidth,height=0.7\linewidth,keepaspectratio]{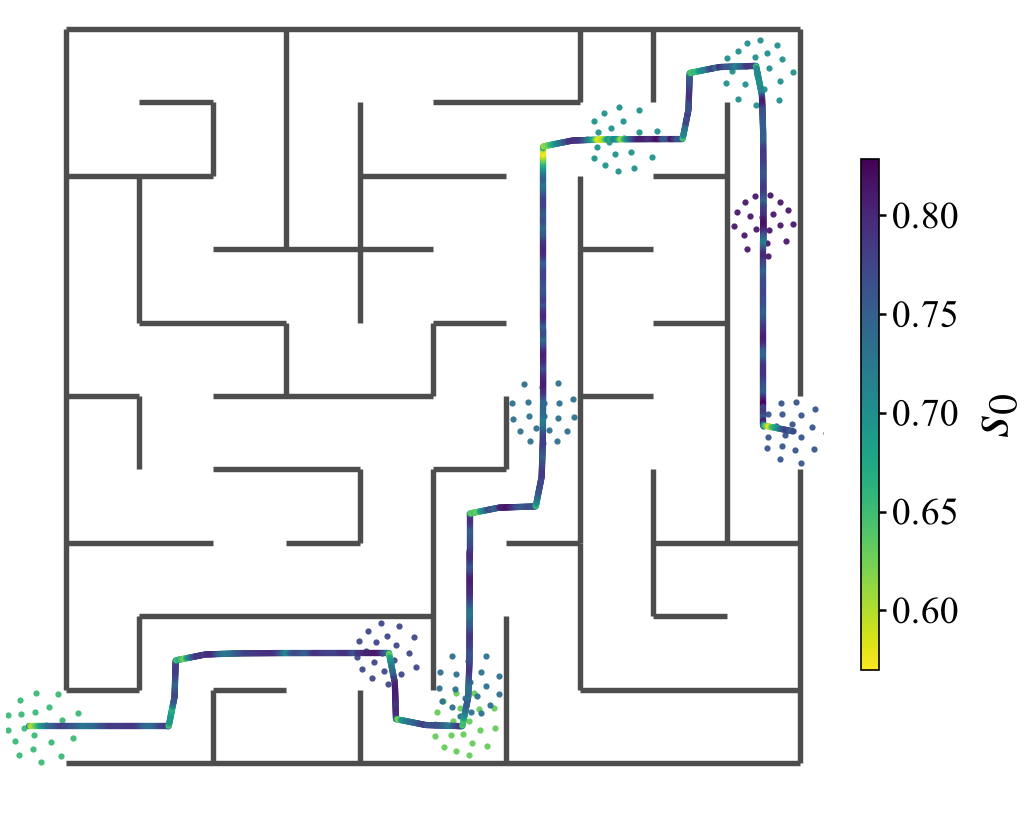}
\vspace{-16px}
\caption{2-D maze traversal with our approach.
The centroid trajectory is colored by the broadcast value $s_0$, dark in narrow corridors and light in open areas, and the route is collision-free.}
\vspace{-8px}
\label{fig:maze2d}
\end{figure}

In both mazes our approach is collision-free and adapts the size of the
collective to the local corridor width: $s_0$ rises and falls along the
route without any schedule, so that the collective fits each passage as
it reaches it (Fig.~\ref{fig:maze2d}). This is the consequence of
placing the size bound in the same QP as the wall constraints; it is
recomputed from the local geometry at every step, so the compression a
passage demands is available the moment the passage is reached. The prior
scheme is collision-free in two dimensions, where its planned
size happens to fit, but collides with the walls and floor tens to over a hundred
times per seed at the three-dimensional partitions, which demand a
compression the plan did not anticipate, some of the collisions passing
through the partitions themselves, whereas our approach remains
collision-free (Fig.~\ref{fig:maze3d}).

\subsection{Multiple collectives}\label{sub:multiF}
\begin{figure*}[t]
\centering
\includegraphics[width=\textwidth,height=0.18\textwidth,keepaspectratio]{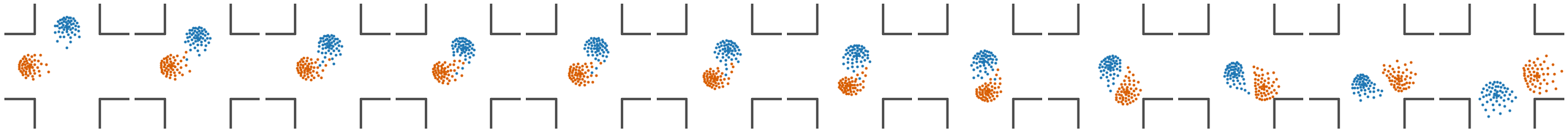}
\caption{Two collectives crossing the junction simultaneously, from approach through closest approach to separation. Neither collective observes the other; the only coordination is a distinct phase label per collective.}
\vspace{-12px}
\label{fig:crossing}
\end{figure*}

\begin{figure}[t]
\centering
\includegraphics[width=\linewidth]{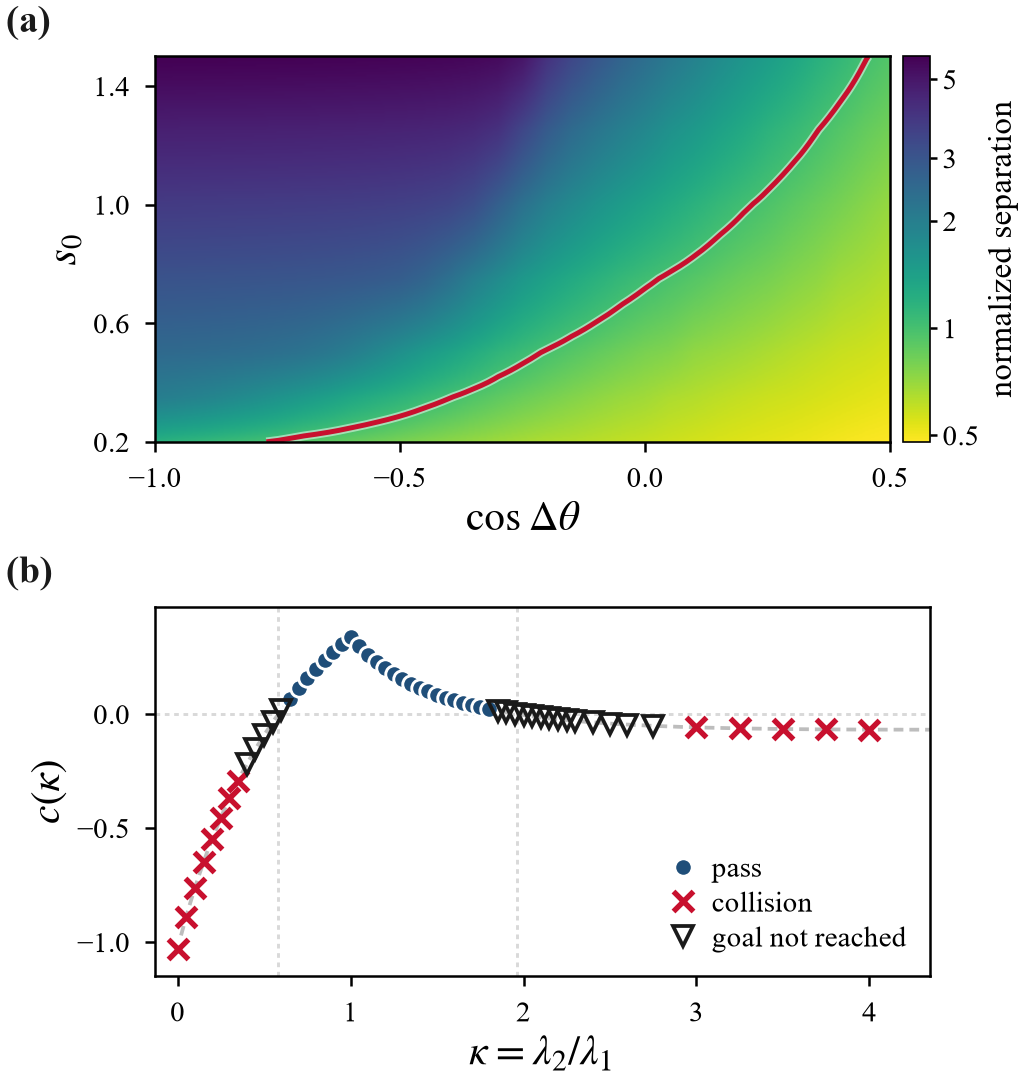}
\vspace{-20px}
\caption{(a) Two collectives placed in contact, over $\cos\Delta\theta$ and the coupling $s_0$. Color is the separation between the collectives normalized by the sum of their radii, and the solid contour is the overlap boundary. (b) Headroom $c(\kappa)$ against the split ratio $\kappa$ (Proposition~\ref{prop:ratio}), with simulated runs marked pass, collision, or goal not reached; vertical lines are the roots of $c(\kappa)$.}
\vspace{-16px}
\label{fig:sweeps}
\end{figure}

When two collectives must pass through the same junction, the usual approach is to let one through while holding the other, which requires knowing where each collective is.
With a distinct fixed phase label assigned to each collective this coordination becomes unnecessary because we can release both collectives at the same time, head on, and they pass without collision while neither observes the other (Fig.~\ref{fig:crossing}).
The runs use a junction width ratio of $3.5$ and a phase difference of $\pi$; a run passes if every collective reaches its goal without collision.
All three seeds pass, and the verdict holds when cohesion (Remark~\ref{rem:cohesion}) is added to the criterion.

The behavior follows from the phase-dependent attraction between the collectives, $A+J\cos\Delta\theta$.
At a phase difference of $\pi$ the coupling term is repulsive, the net force turns repulsive, and the two collectives separate on their own.
A sweep over $\cos\Delta\theta$ and the coupling $s_0$, with no walls present, confirms this: the overlap boundary moves monotonically toward positive $\cos\Delta\theta$ as $s_0$ grows, and at the operating point it lies at $\cos\Delta\theta=-0.39$ (Fig.~\ref{fig:sweeps}(a)).
As $\cos\Delta\theta$ rises, the coupling needed to overcome the net attraction rises with it, and once $\cos\Delta\theta$ becomes positive the coupling term is itself attractive, so the collectives fail to separate and cohesion breaks; in simulation this threshold depends on the coupling.
As the number of collectives grows, Proposition~\ref{prop:phase} forces the worst pair's $\cos\Delta\theta$ up to $\cos(2\pi/M)$, and the required coupling exceeds the ceiling.

The repulsion that separates the collectives is paid for in wall
clearance. At a width ratio of $3.0$ one of three seeds is pushed into a
wall; the minimum wall clearance grows monotonically with the width, and
the failure disappears from ratio $3.5$. Three collectives do not
pass, and handling inter-collective coupling inside the optimization is
the next step.

\section{Limitations}\label{sec:lim}
Three assumptions sit at the level of the model. The support radius of the
cutoff kernel is a design constant chosen by a behavior sweep rather
than a physical quantity, and its correspondence to the effective range of
magnetic interaction is unverified. The assumption that field strength
maps to the coupling coefficient, and the physical realizability of the
ratio $\lambda_2/\lambda_1$, are likewise unverified; the microrobot
literature demonstrates switching collective behavior by field frequency
and amplitude but supplies no map to a coupling coefficient
\cite{ceron2023programmable,gardi2022microrobot,xie2019reconfigurable}.
The model treats robots as points, so beyond roughly $N=117$, where the
nearest-neighbor spacing falls below the robot diameter, the minimum
radius is set by physical packing rather than by the coupling ceiling and
the model no longer applies. There is no hardware validation.

Two conditions underlie the guarantees. First, cohesion is the validity
condition of the reduction (Remark~\ref{rem:cohesion}); the cohesion
constraint bounds the maximum radius but does not by itself guarantee
connectivity, and guaranteeing connectivity structurally requires the
radius bound to be at most half the kernel support. Second, the
theoretical coupling ceiling is $J<B$, but in simulation the collective
fragments from $J/B\approx0.5$; the reason is not identified, our
operating point sits on that boundary, and because the ceiling sets the
minimum passable width directly (Proposition~\ref{prop:wmin}), identifying
the reason is also a question about the performance limit.

Two extensions remain. Operating three or more collectives is, as
Section~\ref{sub:multiF} showed, a matter of handling inter-collective
coupling inside the QP. Further, the guarantee is a continuous-time CBF
condition while the controller runs in discrete steps, and the margin
between them has not been formalized; if the robots carry inertia the
position barrier has relative degree two, so the formulation must change
\cite{xu2026navigation}.

\section{Conclusion}
We have presented a formulation in which a single broadcast input
regulates the motion, deformation, and safety of a robot collective within
one optimization. Prior controllers required a separate command per robot
for motion and deformation and resized the collective by a schedule fixed
at the planning stage, and their single-input variant could only translate
rigidly, so it could not pass a corridor narrower than the collective.
Designing the broadcast as a spatially affine field makes the centroid
and the size of the collective affine in the input, so that both have
relative degree one and the wall and size conditions enter the same QP as
hard constraints. The controller therefore decides the compression at
every step from the walls it sees, the collective passes corridors
narrower than itself, and in the maze the safety constraints hold at every
step.

We have shown that size regulation is a necessity rather than a choice and that a single actuation budget sets both the reachable speed and the minimum passable width, with the pre-registered minimum width verified in simulation to within three percent.
We have further shown that several collectives can be steered individually with distinct phase labels, demonstrated simultaneous passage of two collectives through a junction, and derived the bound beyond which a single label no longer suffices; handling inter-collective coupling inside the optimization is the next step.

\newpage
\section*{Acknowledgment}
\noindent Codex and Claude Code were used to assist with code development and language refinement.

\bibliographystyle{IEEEtran}
\bibliography{refs}

\end{document}